# Anomaly Segmentation for High-Resolution Remote Sensing Images Based on Pixel Descriptors

**Jingtao Li[1], Xinyu Wang[2*], Hengwei Zhao[1], Shaoyu Wang[1], Yanfei Zhong[1]**

[1] State Key Laboratory of Information Engineering in Surveying, Mapping and Remote Sensing, Wuhan University, P. R. China
[2] School of Remote Sensing and Information Engineering, Wuhan University, P. R. China
{JingtaoLi, wangxinyu, whu_zhaohw, wangshaoyu, zhongyanfei}@whu.edu.cn

## Abstract

Anomaly segmentation in high spatial resolution (HSR) remote sensing imagery is aimed at segmenting anomaly patterns of the earth deviating from normal patterns, which plays an important role in various Earth vision applications. However, it is a challenging task due to the complex distribution and the irregular shapes of objects, and the lack of abnormal samples. To tackle these problems, an anomaly segmentation model based on pixel descriptors (ASD) is proposed for anomaly segmentation in HSR imagery. Specifically, deep one-class classification is introduced for anomaly segmentation in the feature space with discriminative pixel descriptors. The ASD model incorporates the data argument for generating virtual abnormal samples, which can force the pixel descriptors to be *compact* for normal data and meanwhile to be *diverse* to avoid the model collapse problems when only positive samples participated in the training. In addition, the ASD introduced a multi-level and multi-scale feature extraction strategy for learning the low-level and semantic information to make the pixel descriptors *feature-rich*. The proposed ASD model was validated using four HSR datasets and compared with the recent state-of-the-art models, showing its potential value in Earth vision applications.

## Introduction

Anomaly segmentation is aimed at segmenting the anomaly patterns which deviate from the normal patterns (Pimentel et al. 2014; Pang et al. 2021). Due to the lack of abnormal samples, anomaly segmentation is a challenging task, but plays an important role in many computer vision applications, including medical analysis (Fernando et al. 2021), industrial defect detection (Bergmann et al. 2019), video surveillance (Liu, Li, and Poczos 2018), and environmental monitoring (Miau and Hung 2020).

Anomaly segmentation in high spatial resolution (HSR) remote sensing images (e.g., Figure 1) is a powerful tool for environmental monitoring (Miau and Hung 2020; Wang et al. 2019). Despite this, few related works have focused on anomaly segmentation in HSR imagery because of the unique characteristics when compared to the industrial and medical images used in most anomaly segmentation tasks, which have a regular structure. The objects in HSR images typically have a more complex spatial distribution and large radiation differences within the same class. Furthermore, since HSR images can be captured in different angles and heights, the objects always have multiple scales and show rotation invariance. These characteristics make anomaly segmentation for HSR images a challenging task.

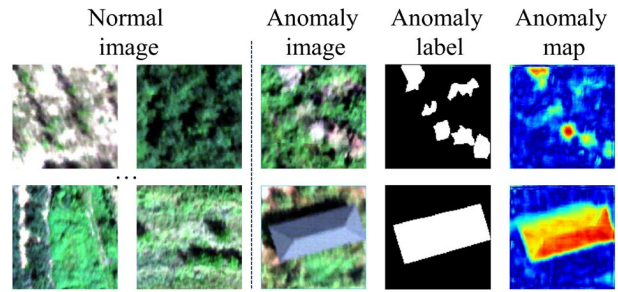

Figure 1: Anomaly segmentation example for HSR remote sensing images using proposed model. In the forest scene, the common forest pattern is considered as normal, and the abnormal objects such as diseased trees (the first row) and the house in the forest (the second row) are identified as anomalies.

The mainstream anomaly segmentation models detect anomalies in the image space, where the anomaly score is computed based on the image pixel values. Typical examples are the autoencoder (AE)-based models (Zavrtanik, Kristan, and Skočaj 2021; Gong et al. 2019) and the generative adversarial network (GAN)-based models (Ngoetal. 2019; Zenatietal. 2018b). AE-based models assume that normal samples can be reconstructed more easily than the anomalous ones and the reconstruction error indicates the anomaly segmentation score (Pang et al. 2021). However, the low-level reconstruction error has been shown to focus

---

on the pixel-wise error, resulting in abnormal samples also being reconstructed, especially when the normal distribution is complex. (Fei et al. 2020; Gong et al. 2019; Zong et al. 2018). GAN-based models detect anomalies from the generation performance (Akcay, Atapour-Abarghouei, and Breckon 2018; Ngo et al. 2019; Xia et al. 2022), where the superior capability in generating image data also empowers the detection of abnormal samples (Pang et al. 2021). In spite of this, the complex distribution of HSR images can make the generator generate data instances that are out of the manifold of normal instances (Pang et al. 2021).

Differing from the AE-based and GAN-based models, deep one-class classification (OCC)-based methods detect the anomalies in the feature space (Shi, Yang, and Qi 2021; Lei et al. 2021; Zhao et al. 2022; Li et al. 2022), where the anomaly score is computed based on the extracted image descriptors. These methods aim to learn discriminative descriptors in the training stage and compute the anomaly score in the feature space using a measurement such as the Mahalanobis or Euclidean distance (Reiss et al. 2021; Ruff et al. 2018; Shi, Yang, and Qi 2021). Because deep OCC-based methods focus on semantic features rather than low-level pixel errors, it is more suitable to deal with the anomaly segmentation task in HSR imagery which has complex distribution. However, two barriers exist when applying existing methods directly. (i) Due to the lack of abnormal samples, the model training only uses normal samples and is optimized to be compact (Ruff et al. 2018; Chalapathy, Menon, and Chawla 2018), which can easily result in the model collapse problem (Reiss et al. 2021). (ii) The anomalies in HSR imagery have rich low-level (e.g., texture) and high-level (e.g., semantic) features, which are both important for the anomaly segmentation task and real application. Although the current deep OCC models can capture useful semantic features, they perform suboptimally than models detecting in the image space for samples with regular structures (Li et al. 2021), because low-level features are mostly forgotten in feature space.

In this paper, we tackle the two problems for the anomaly segmentation task using HSR images. A novel anomaly segmentation model based on pixel descriptors (ASD) is proposed. (i) In addition to considering the *compact* property of the obtained descriptors, the ASD model encourages descriptors to be *diverse* by increasing the descriptor distance between the original image and the transformed image with the use of data augmentation techniques. The transformed descriptors act as anomalies, to some extent, which enhances the anomaly detection ability and prevents simultaneous model collapse. (ii) To make the descriptor *feature-rich*, a descriptor at different scales is fused for each pixel, and an auxiliary reconstruction head is designed to force the descriptor to remember the low-level features. Compact, diverse, and feature-rich property optimizes the model together from the perspective of the feature distance and feature quantity. ASD sets the first baseline for the anomaly segmentation task in HSR imagery.

The ASD model was validated on four HSR datasets: the DeepGlobe land-cover segmentation dataset, the Agriculture-Vision agriculture pattern segmentation dataset, the Landslide4Sense landslide detection dataset, and the forest anomaly detection dataset (FAS, made by ourselves). The ASD model showed an obvious superiority over the recent state-of-the-art anomaly segmentation models (with an area under the curve (AUC) improvement of 5–10 points in most cases). The results obtained on the Landslide4Sense and FAS datasets confirmed the great application potential of the ASD model in disaster detection and forest monitoring.

## Related Work

**AE-based models** are always composed of an encoding and decoding network, with the aim being to reconstruct the original input data (Pimentel et al. 2014). Hawkins et al. (2022) first introduced the AE into the anomaly detection field, where the features learned in the latent space can be used to distinguish normal and anomalous data. The reconstruction error is considered as the anomaly degree and the mean square error (MSE) is adopted as the loss function in most studies (Pang et al. 2021). To promote the performance, Pathak et al. (2016) blanked the input image randomly and forced the model to reconstruct the damaged area. Similarly, the ARNet model was proposed, which erases some input attributes and reformulates the problem as a restoration task (Fei et al. 2020). Recently, Zavrtanik et al. (2021) cast the reconstruction problem as an inpainting problem and reconstructed the image from partial inpaintings. However, the extracted low-level features can be shared by both normal and anomalous data (Fei et al. 2020) when dealing with complex HSR images.

**GAN-based models** aim to generate the image rather than reconstruct it. As one of the early GAN-based models, the AnoGAN model assumes that the learned latent space can represent normal samples well, but not the anomalous samples (Schlegl et al. 2017). Given a test image, the difference between the regenerated image obtained using the searched latent feature and the test image is considered as the anomaly degree. The famous GANomaly model improved the generator architecture from a decoder to an encoder-decoder encoder design and used high-level features to assist computing the anomaly score (Akcay, Atapour-Abarghouei, and Breckon 2018). GAN-based models have demonstrated superior capabilities in generating image data, which also empowers the detection of abnormal samples (Pang et al. 2021). In spite of this, the complex distribution of HSR images can make the generator generate data instances that are out of the manifold of normal instances (Pang et al. 2021).

**One-class classification models** are also used in some anomaly segmentation works (Pang et al. 2021). One of their greatest advantages over the AE-based and GAN-based models is that the OCC models detect anomalies in the feature space with high-level semantic information. They first divide an image into many patches and then learn the corresponding representations. The anomaly score is computed in the feature space using a measurement such as the Mahalanobis or Euclidean distance (Reiss et al. 2021; Ruff et al. 2018; Shi, Yang, and Qi 2021). Most OCC models are based on the principle of one-class support vector machine (OCSVM) (Sch¨olkopf et al. 1999; Andrews, Morton, and Griffin 2016) or support vector data description (SVDD) (Tax and Duin 1999; Chalapathy, Menon, and Chawla 2018; Ruff et al.2018). However, these models mainly consider the compact property of the obtained one class features, resulting in the model collapse problem (Reiss et al. 2021), and they lack consideration of the low-level structural features.

# Methodology

*Overview.* This section describes the core principles of the proposed ASD model. The overall workflow of the ASD model is shown in Section 3.1, which includes two steps: descriptor extracting and anomaly score computation. To extract the ideal descriptors, descriptor learning is the most important part and is described detailed in Section 3.2. The computation method of the anomaly score is given in Section 3.3.

## Overall Workflow of The ASD Model

Given an HSR image $X$ with size $H \times W \times B$, where $H$, $W$, and $B$ are the height, width and bands of the image, the anomaly segmentation task can be viewed as a mapping function $f$ from the $X$ to the anomaly map $A$ with size $H \times W$. Each pixel in the anomaly map is in the range [0,1]. Generally speaking, the higher the value in the anomaly map, the higher the anomaly degree.

The ASD model separates the function $f$ into two steps and the overall workflow is shown in Figure 2. The first step $f_1$ extracts the dense descriptor cube $D$ for each image pixel, which is the core part and also the training focus in the ASD model. The descriptors are expected to contain important visual characteristics for the anomaly segmentation task. To incorporate the pixel context and obtain fine pixel correspondence, the patch-based paradigm is chosen to compute the descriptor $F$ for the center pixel $x$. In this step, the $H \times W$ patches form the input samples and a descriptor cube $D$ with size $H \times W \times L$ is output, where $L$ is the descriptor length.

The second step $f_2$ outputs the anomaly map based on the trained descriptor encoders in the first step. Specifically, the trained descriptors of the training samples are modeled as a multivariate Gaussian Distribution (MGD) (Guimaraes et al. 2018) by the Gaussian Density Estimate (GDE). For the test descriptor, its Mahalanobis distance from the MGD is used to measure the anomaly score. The formal mapping of $f$, $f_1$, and $f_2$ is shown in Eqs. (1-3).

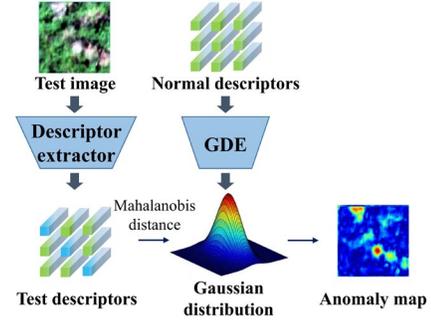

Figure 2: The overall workflow of the ASD model, which includes two steps. In the first step, the ASD model extracts a descriptor for each pixel with the descriptor extractor. In the second step, the descriptors for normal scenes are modeled as the Gaussian distribution, and the Mahalanobis distance between the test descriptor and the modeled distribution is considered to measure the anomaly score.

$$f : X \to A \quad (1)$$
$$f_1 : X \to D \quad (2)$$
$$f_2 : D \to A \quad (3)$$

## Ideal Descriptors Learning

The descriptors obtained in the first step (as mentioned in Section 3.1) are expected to contain important visual characteristics for the anomaly segmentation task. To achieve this aim, ideal descriptors are optimized using three conditions from the characteristics of the anomaly segmentation task and HSR images.

**Compact.** One of the characteristics of anomaly segmentation is that only normal samples are used in the training stage. In other words, all the training samples are of the same class, which naturally results in compact visual descriptors in the feature space. This compactness is also a useful supervised signal for the anomaly segmentation task.

To keep $D$ compact, an enclosing hypersphere around all the pixel descriptors is constructed, which is motivated by the deep SVDD method (Ruff et al. 2018). We let $R$ be the hypersphere radius and $C$ be the center. The $L_1$ loss (Eq. (4)) aims to minimize the hypersphere radius and the distance from the obtained pixel descriptors to the center $C$, where the parameter $\lambda$ controls the trade-off between the size of the hypersphere and the number of surrounded descriptors. The maximum distance between $C$ and $F$ in $D$ is chosen to compute the radius $R$. Compared to using the mean value, this setting helps the model focus on special normal samples, rather than just considering them as noise.

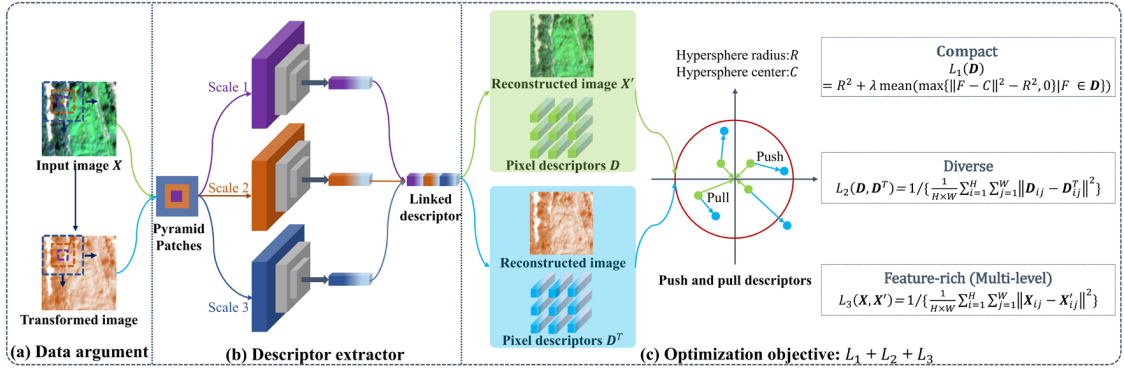

Figure 3: The descriptor optimization process of the ASD model. (a) For each normal image, its transformed image is generated using data argumentation techniques for generating the artificially negative samples. (b) The ASD model is designed as a two-head architecture. One head outputs the dense descriptor and the other reconstruction head is designed to force the obtained descriptors to contain both *high-level* and *low-level* features. Pyramid patches are extracted at different scales for the *multi-scale* features. (c) To obtain the *ideal descriptors*, as defined in Section 3.2, the optimization tries to find a *compact* hypersphere surrounding all the descriptors of the original image by pulling them to the center, keeping the descriptors *diverse* by increasing the distance between the original descriptors and the transformed descriptors.

$$L_1(D) = R^2 + \lambda \operatorname{mean}(\max\{\|F - C\|^2 - R^2, 0\} | F \in D) \quad (4)$$

**Diverse.** Compactness is the first basic condition. However, the model can easily collapse if only a compactness constraint used. In other words, the model would map all the input samples into the same point. This "cheating" makes the model lose the anomaly detection ability. To deal with this problem, the diverse condition is necessary, which stresses that a different pixel $x$ obtains different values of $F$.

The key consideration to keeping the descriptors diverse is to keep the model sensitive to the input sample change. Considering the fact that training images are always anomaly free and real negative samples are difficult to obtain, data augmentation techniques, such as the channel shuffle operation, are used to generate negative samples. Formally, the augmentation operation set $S_a = \{A_1, A_2, ..., A_n\}$ contains $n$ kinds of different augmentation operations. For the original image $X$, the obtained image descriptor cube $D$ can be seen as a positive one. Then, after applying the operations from $S_a$ on $X$ in turn, $X^T$ can be obtained and the corresponding cube $D^T$ is considered to be a negative sample. Eqs. (5-6) formally express the above process.

$$X^T = A_n(\ldots(A_2(A_1(X^T)))) \quad (5)$$
$$D = f_1(X), \quad D^T = f_1(X^T) \quad (6)$$

Both $D$ and $D^T$ have the same shape $H \times W \times L$. The diversity loss is defined as the average pixel descriptor difference between $D$ and $D^T$, as shown in Eq. (7). With the $L_2$ loss, the model is encouraged to increase the sensitivity to the input difference.

$$L_2(D, D^T) = 1 / \left\{ \frac{1}{H \times W} \sum_{i=1}^{H} \sum_{j=1}^{W} \left\| D_{ij} - D_{ij}^T \right\|^2 \right\} \quad (7)$$

Some technologies have the potential to deal with the model collapse, such as reducing the model bias (Ruff et al. 2018) or designing early-stopping strategies (Reiss et al. 2021). However, the proposed $L_2$ loss does not need early-stopping or change of the model architecture.

**Feature-rich.** The compact and diverse conditions measure the descriptors from the perspective of distance. The feature-rich condition measures the descriptors from the perspective of the amount of representative information. In the ASD model, multi-scale and multi-level features are considered in particular.

The multi-scale characteristic is an import difference for HSR images, compared to natural images. For example, large-scale information is important for rivers and small-scale information is important for urban buildings. Even for the same scene, the images are always taken at different heights, which poses a challenge for the model ability to catch the multi-scale information.

To enhance the model ability to deal with multi-scale information, the ASD model uses a resize operation set $S_s = \{U_1, U_2, ..., U_m\}$ for the input patches. Given an image $X$, it is resized using each operation $U_i$ in $S_s$, and obtains $m$ different-scale versions $X_1, X_2, ..., X_m$ of the same image. Then, for each center pixel $x$, $m$ patches are cropped with size $P \times P$ from the $m$ scaled images. Next, the obtained pyramid patches are fed into with $m$ individual encoders $E_1, E_2, ..., E_m$, and $m$ pixel descriptors are obtained, where each descriptor has the same length $L$. The $m$ pixel descriptors are then concatenated further to form a descriptor vector with length $m \times L$. The descriptor cube $\boldsymbol{D}_c$ with size $H \times W \times (m \times L)$ is naturally obtained when all the pixels in $\boldsymbol{X}$ are processed. Finally, a $1 \times 1$ convolution operation is used to map the concatenated descriptors into size $L$. This is the process for extracting the final pixel descriptors. Eqs. (8-10) formally express the above process, which is also the detailed process of $f_1$. Figure 3 shows the process when $m = 3$.

$$\boldsymbol{X}_1, \boldsymbol{X}_2, ..., \boldsymbol{X}_m = U_1(\boldsymbol{X}), U_2(\boldsymbol{X}), ..., U_m(\boldsymbol{X}) \quad (8)$$

$$\boldsymbol{D}_c = \text{concat}([M(\boldsymbol{X}_1, \boldsymbol{X}_2, ..., \boldsymbol{X}_m)]) \quad (9)$$

$$\boldsymbol{D} = \text{Conv}_{1 \times 1}(\boldsymbol{D}_c) \quad (10)$$

Multi-level features are necessary when dealing with the various objects in the anomaly segmentation task. Although the deep architecture extracts high-level semantic information through the descriptors, the low-level information such as texture is gradually forgotten as the network goes deeper. This is beneficial for objects such as buildings, but is not expected for some objects such as water and river because the texture feature is useful for them.

To ensure that both high-level and low-level features are contained in the descriptors, the ASD model is designed as a two head architecture. Both heads grow from the concatenated descriptor cube $\boldsymbol{D}_c$. One head uses the $1 \times 1$ convolution operation to obtain the final pixel descriptors. The other head also uses the $1 \times 1$ convolution but aims to reconstruct the original pixel. To reconstruct the pixel value, the concatenated descriptors are forced to contain the low-level features. Note that the reconstruction head is only used in the training and is abandoned in the test stage. $\boldsymbol{X}'$ denotes the reconstructed image, and the MSE is used to compute the loss (Eqs. (11-12)).

$$\boldsymbol{X}' = \text{Conv}_{1 \times 1}(\boldsymbol{D}_c) \quad (11)$$

$$L_3(\boldsymbol{X}, \boldsymbol{X}') = \frac{1}{H \times W} \sum_{i=1}^{H} \sum_{j=1}^{W} \left\| \boldsymbol{X}_{ij} - \boldsymbol{X}'_{ij} \right\|^2 \quad (12)$$

In total, the three properties: compact, diverse and feature-rich work together to design the model architecture and optimize the descriptor learning. The optimization objective of the ASD model is the sum of the above losses, as shown in Eq. (13). Figure 3 shows the overall descriptor learning process.

$$Loss = L_1(\boldsymbol{D}) + L_2(\boldsymbol{D}, \boldsymbol{D}^T) + L_3(\boldsymbol{X}, \boldsymbol{X}') \quad (13)$$

## Anomaly Score Computation

When the descriptor optimization process of step $f_1$ is finished, the second step $f_2$ outputs the anomaly map based on the optimized descriptors. There exist various methods to complete step $f_2$. Although non-parametric statistical methods do not rely on any distribution assumption, it requires a lot of samples to achieve accurate estimation and can be computationally expensive e (Pang et al. 2021). Conversely, parametric density estimation needs fewer samples, and the Gaussian assumption holds in most cases (Pimentel et al. 2014).

In the ASD model, the Gaussian assumption is adopted to model the normal descriptors. Using the normal samples in the training stage, the mean $\mu$ and the covariance matrix $\boldsymbol{\Sigma}$ can be estimated. Given a test descriptor $x_t$, its Mahalanobis distance from the modeled distribution (as shown in Eq. (14)) is considered the anomaly degree, which can be converted to the anomaly score after the normalization.

$$Anomaly\ degree = \sqrt{(x_t - \mu)^T \boldsymbol{\Sigma}^{-1}(x_t - \mu)} \quad (14)$$

## Experiments

### Experimental Settings

**Datasets**
The proposed ASD model was evaluated on four HSR image datasets: DeepGlobe (Demir et al. 2018), Agriculture-Vision (Chiu et al. 2020), FAS, and Landslide4Sense (Ghorbanzadeh et al. 2022). The DeepGlobe and Agriculture-Vision datasets were originally made for the land-cover segmentation and agriculture pattern segmentation tasks, respectively. To adapt these datasets for the anomaly segmentation task, the pixels of the remaining classes were masked for a fixed normal class in the training process to keep the anomaly-free characteristic.

To show the application value of the ASD model, the FAS and Landslide4Sense datasets were used. The FAS dataset was made by ourselves for the forest monitoring application, where the common forest pattern (i.e., Figure 1) is treated as the normal class, and some abnormal objects, such as house, lake, car, and diseased tree, are considered as anomalies. The RGB imagery in the FAS dataset was made from UAV-borne hyperspectral images in forest scene. The pixel resolution is 11 cm and the image size is 120×120. In the Landslide4Sense dataset, the anomaly segmentation model was used to segment the landslide area by learning from the normal mountain pattern.

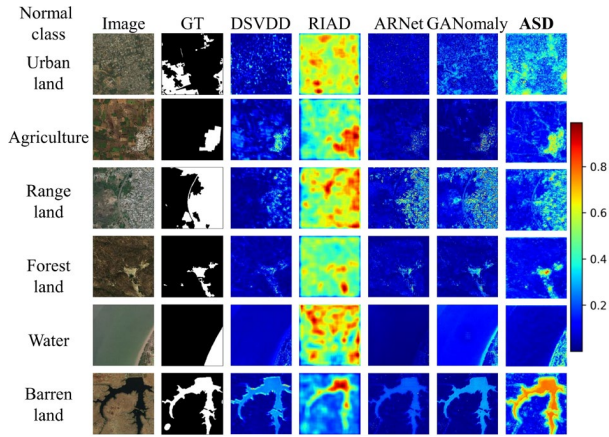

Figure 4: The anomaly segmentation results obtained on the DeepGlobe dataset for each normal class, White pixels cover the anomalous region.

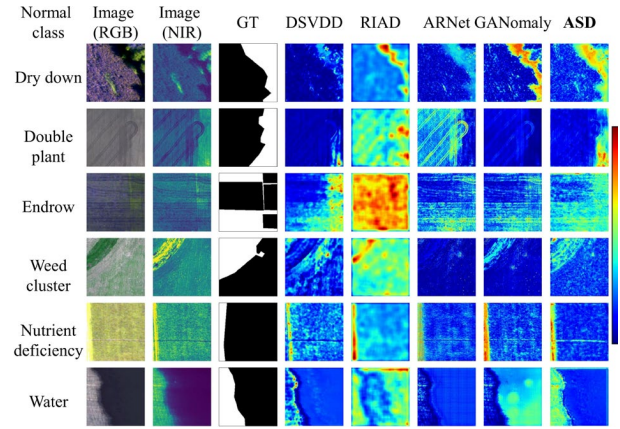

Figure 5: The anomaly segmentation results obtained on the Agriculture-Vision dataset for the six normal classes.

**Comparative Models and Evaluation Metrics**

The ASD model was compared with four state-of-the-art methods covering both image space and feature space types. These methods include GANomaly (Akcay, Atapour-Abarghouei, and Breckon 2018), ARNet (Fei et al. 2020), RIAD (Zavrtanik, Kristan, and Skocaj 2021) and deep SVDD (DSVDD) (Ruff et al. 2018). For the GANomaly, RIAD, and DSVDD, the model hyper-parameters were kept same as the authors' open source code. ARNet was implemented using the same architecture as RIAD. The model performance was evaluated using the area under the curve (AUC) metric and the mean Intersection over Union (mIOU). The segmentation threshold for the mIOU corresponds to the left-upper point of the Receiver operating characteristic (ROC) curve.

**Implementation Details**

The fast version (Bailer et al. 2018) of the point descriptor extraction network in the work of Simo-Serra et al. (2015) acted as the pixel descriptor encoder in the proposed model. In all the experiments, the models were trained for 100 epochs, and the batch size was 1. The Adam optimizer with learning rate 0.0001 was used. $\lambda$ was set to 10. $S_s$ was set to $\{0.5, 1.0, 2.0\}$ and $P$ is 15 for all the descriptor encoders. The first 10 epochs were trained using only the $L_3$ loss to compute the initial $C$. $R$ was initialized to 3.0. $C$ and $R$ were updated after each epoch using all the training descriptors. The dimension $L$ was set to 5. The data augmentation operations used in the ASD and ARNet model were the GaussNoise, ChannelShuffle, RandomBrightness, RandomContrast, and Solarize operations (implemented with the Albumentations tool (Buslaev et al. 2020)). Due to the AUC computation burden, 2000 test images in Agriculture-Vision dataset were chosen to be evaluated. The CPU was an Intel(R) Xeon(R) CPU E5-2690 v4 @ 2.60 GHz with 62.6 GB memory, and the GPU was a Tesla P100-PCIE with 16 GB of memory.

| Method | Urban land | | Agriculture | | Range land | | Forest land | | Water | | Barren land | |
|---|---|---|---|---|---|---|---|---|---|---|---|---|
| | AUC | mIOU | AUC | mIOU | AUC | mIOU | AUC | mIOU | AUC | mIOU | AUC | mIOU |
| DSVDD | 57.0 | 31.5 | 60.3 | 41.4 | 53.6 | 16.3 | 58.7 | 24.0 | 37.6 | 2.2 | 50.6 | 17.7 |
| RIAD | 52.3 | 12.4 | **65.9** | **46.3** | 47.6 | 7.0 | 69.4 | 33.1 | 57.7 | 43.0 | 53.3 | 13.9 |
| ARNet | 50.2 | 39.0 | 60.1 | 40.9 | 48.2 | 6.9 | 67.6 | 34.5 | 54.7 | 12.5 | 61.4 | 33.3 |
| GANomaly | 42.8 | **44.3** | 51.7 | 35.8 | **55.1** | **30.6** | 75.4 | 41.3 | 58.8 | **44.5** | 36.8 | **39.7** |
| ASD | **63.4** | 38.5 | 64.1 | 42.7 | 54.3 | 23.2 | **79.5** | **43.1** | **73.3** | 39.5 | **62.5** | 34.9 |

Table 1: The anomaly segmentation results obtained on the DeepGlobe dataset.

| Method | Drydown | | Double plant | | Endrow | | Weed cluster | | ND | | Water | |
|---|---|---|---|---|---|---|---|---|---|---|---|---|
| | AUC | mIOU | AUC | mIOU | AUC | mIOU | AUC | mIOU | AUC | mIOU | AUC | mIOU |
| DSVDD | 60.9 | 30.8 | 53.0 | 14.7 | 54.6 | 10.1 | 48.0 | 3.67 | 59.6 | 24.4 | 72.1 | 26.0 |
| RIAD | 62.2 | 31.0 | 60.9 | **25.6** | 59.3 | **27.7** | 55.2 | 38.2 | 63.8 | 25.7 | 86.1 | **42.7** |
| ARNet | 61.1 | 30.6 | 51.5 | 15.3 | 57.1 | 25.5 | 53.0 | 15.9 | 59.9 | 26.8 | 45.3 | 9.6 |
| GANomaly | 59.5 | 26.3 | 49.6 | 4.2 | 56.5 | 26.4 | 51.1 | **41.9** | 62.9 | **33.1** | 64.2 | 20.7 |
| ASD | **67.4** | **36.4** | **61.3** | 24.8 | **61.1** | 25.7 | **58.0** | 19.7 | **65.9** | 31.7 | **90.0** | 40.4 |

Table 2: The comparative quantitative anomaly segmentation results on the Agriculture-Vision dataset. (ND is nutrient deficiency)

**Results on the DeepGlobe Dataset**

The quantitative and qualitative results are reported in Table 1 and Figure 4, respectively. In Table 1, the ASD model achieves the highest AUC values for the four normal classes. For the Urban land class, the ASD model surpasses the second-best model by over 6 points, showing its superiority when dealing with a complex distribution. In Figure 4, the anomaly maps obtained by the ASD model are the closest to the ground truth.

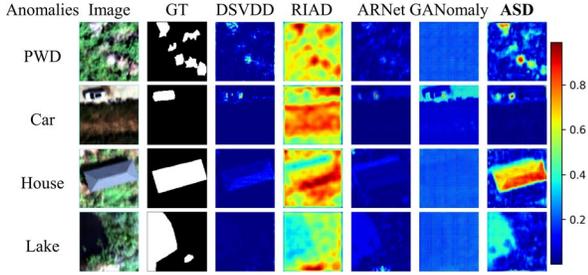

Figure 6: The anomaly segmentation results obtained on the FAS dataset. The common forest pattern (see Figure 1) is considered as normal, and four anomalies are considered.

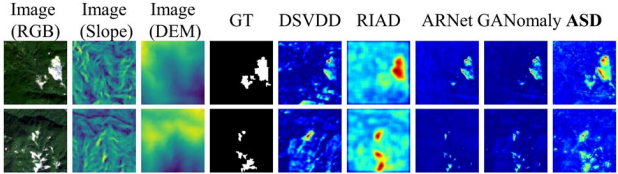

Figure 7: The anomaly segmentation results obtained on the Landslide4Sense dataset. The common mountain pattern is considered as normal, and the landslides are the anomalies.

### Results on the Agriculture-Vison Dataset

Table 2 and Figure 4 respectively show the quantitative and qualitative results for the Agriculture-Vision dataset. In Table 2, the ASD model achieves the best AUC results for all six normal classes. ASD surpasses the second-best model by 5 points for the Drydown class. Except Weed cluster, the mIOU values of ASD are all close to the optimal value. In Figure 5, it can be seen that accurate results and fine boundaries are obtained by the ASD model for most classes. For the normal class of water, only the ASD model outputs a correct anomaly map, and some models completely reverse the anomaly regions.

### Results on the FAS and Landslide4Sense Datasets

The FAS and Landslide4Sense datasets were used to show the application value of the proposed anomaly segmentation model in forest monitoring and landslide detection. Table 3, Figure 6, and Figure 7 report the related results. In both datasets, the ASD model achieves the best AUC and mIOU scores. Satisfactory anomaly maps are obtained, demonstrating great application value.

### Ablation Studies

The core idea of the ASD model is to find ideal descriptors, so three loss constraints corresponding to the conditions described in Section 3.2 were designed. Table 4 illustrates the effectiveness of three losses for different types of earth vision scenes $L_1$ (compact loss) can better handle the scene with simple spatial distribution, i.e., Agriculture, Forestland, and Water; (from the first 3 rows). $L_3$ (feature-rich loss) works on the complex scenes, i.e., Urban land and Barren land; (from the 3,5 and 6 rows). $L_2$ (diversity loss) aims at further improving segmentation performance by artificial anomaly samples. (Comparing rows 1 and 3 with 4 and 5, respectively).

| Dataset | DSVDD | | RIAD | | ARNet | | GANomaly | | ASD | |
|---|---|---|---|---|---|---|---|---|---|---|
| | AUC | mIOU | AUC | mIOU | AUC | mIOU | AUC | mIOU | AUC | mIOU |
| FAS | 74.1 | 46.2 | 44.3 | 36.5 | 82.7 | 52.9 | 50.7 | 24.9 | **91.0** | **69.3** |
| Lanslide4Sense | 61.6 | 20.7 | 83.7 | 41.0 | 78.8 | 48.7 | 82.2 | 39.1 | **89.8** | **49.3** |

Table 3: The anomaly segmentation results obtained on the FAS and Landslide4Sense datasets.

| Constraints | Urban land | | Agriculture | | Range land | | Forest land | | Water | | Barren land | |
|---|---|---|---|---|---|---|---|---|---|---|---|---|
| | AUC | mIOU | AUC | mIOU | AUC | mIOU | AUC | mIOU | AUC | mIOU | AUC | mIOU |
| $L_1$ | 51.3 | 38.9 | 62.5 | 42.3 | 52.8 | 18.1 | 76.4 | 42.0 | 70.2 | 35.1 | 56.6 | 26.9 |
| $L_2$ | 40.9 | **45.3** | 59.4 | 37.9 | 53.7 | **40.2** | 76.7 | 47.6 | 71.5 | 36.1 | 49.0 | 34.1 |
| $L_3$ | 62.5 | 35.9 | 60.4 | 38.6 | 52.9 | 19.6 | 75.6 | 41.5 | 68.8 | 36.0 | 61.2 | 32.9 |
| $L_1+L_2$ | 56.5 | 32.4 | 61.0 | 39.7 | **54.7** | 31.8 | 78.8 | **48.6** | 72.5 | 40.6 | 54.6 | 22.8 |
| $L_2+L_3$ | **64.5** | 45.1 | 62.0 | 42.5 | 54.6 | 22.1 | 77.4 | 43.1 | 73.0 | **41.7** | 54.3 | 23.8 |
| $L_1+L_3$ | 64.1 | 45.1 | 62.7 | 41.9 | 52.9 | 20.0 | 77.3 | 43.1 | **74.5** | 41.5 | 61.7 | 31.5 |
| $L_1+L_2+L_3$ | 63.4 | 38.5 | **64.1** | **42.7** | 54.3 | 23.2 | **79.5** | 43.1 | 73.3 | 39.5 | **62.5** | **34.9** |

Table 4: The ASD model ablation analysis for the three loss constraints on the anomaly segmentation results obtained using the DeepGlobe dataset.

### Conclusion

In this paper, we have proposed a pixel descriptor based model for the anomaly segmentation task in HSR imagery. The core innovations are: 1) The three conditions that the ideal descriptor should meet are given from the characteristics of the anomaly segmentation task and HSR images. 2) The corresponding constraints and architecture were designed on this basis. Obvious improvement was achieved on four datasets (including real anomalies in forest and mountain scenes). Overall, proposed model sets the first baseline for the anomaly segmentation task of complex HSR imagery.

### Acknowledgments

This work was supported by National Natural Science Foundation of China under Grant No.42071350 and No.42101327, in part by the Fundamental Research Funds for the Central Universities under Grant 2042021kf0070, and LIESMARS Special Research Funding.